\documentclass[11pt]{article}

\usepackage[utf8]{inputenc}
\usepackage[T1]{fontenc}
\usepackage{lmodern}

\usepackage{graphicx}%
\usepackage{multirow}%
\usepackage{amsmath,amssymb,amsfonts}%
\usepackage{mathtools}
\usepackage{amsthm}%
\usepackage{mathrsfs}%
\usepackage[title]{appendix}%
\usepackage{xcolor}%
\usepackage{textcomp}%
\usepackage{manyfoot}%
\usepackage{booktabs}%
\usepackage{algorithm}%
\usepackage{algorithmicx}%
\usepackage{algpseudocode}%
\usepackage{listings}%
\usepackage[acronym]{glossaries}
\usepackage{caption}
\usepackage{subcaption}
\usepackage{booktabs}
\usepackage[english]{babel}
\usepackage{microtype}

\newcommand{\ignore}[1]{}
\newacronym{gd}{GD}{Gradient Descent}
\newacronym{ot}{OT}{Optimal Transport}
\newacronym{dr}{DR}{Dimensionality Reduction}
\newacronym{gw}{GW}{Gromov-Wasserstein}
\newacronym{mds}{MDS}{Multidimensional Scaling}
\newacronym{isomap}{Isomap}{Isometric Mapping}
\newacronym{gw-mds}{GW-MDS}{Gromov-Wasserstein MDS}
\newacronym{pca}{PCA}{Principal Component Analysis}
\newacronym{ewca}{EWCA}{Entropic Wasserstein Component Analysis}
\newacronym{srgw}{srGW}{Semi-Relaxed Gromov-Wasserstein}
\newacronym{srgw-mds}{srGW-MDS}{Semi-Relaxed Gromov-Wasserstein MDS}
\newacronym{kl}{KL}{Kurdyka–Łojasiewicz}

\renewcommand{\min}[1]{\underset{#1}{\text{min}}\,}

\newcommand{\argmin}[1]{\underset{#1}{\text{argmin}}\,}

\theoremstyle{thmstyleone}%

\theoremstyle{thmstyletwo}%

\theoremstyle{thmstylethree}%
\newtheorem{definition}{Definition}%

\usepackage{amssymb}
\usepackage{amsmath}

\title{Structure-Preserving Multi-View Embedding Using Gromov-Wasserstein Optimal Transport}

\author{
Rafael Pereira Eufrazio$^{1,3}$ \\
\texttt{rafael.eufrazio@ifce.edu.br}
\and
Eduardo Fernandes Montesuma$^{2}$ \\
\texttt{eduardo.montesuma@sigmanova.ai}
\and
Charles Casimiro Cavalcante$^{3}$ \\
\texttt{charles@ufc.br}
}

\date{
\small
$^{1}$Instituto Federal de Educação, Ciência e Tecnologia do Ceará, Canindé, Ceará, Brazil\\
$^{2}$Sigma Nova, Paris, France\\
$^{3}$Department of Teleinformatics Engineering, Federal University of Ceará, Fortaleza, Ceará, Brazil
}

\begin{document}


\maketitle





\begin{abstract}
Multi-view data analysis seeks to integrate multiple representations of the same samples in order to recover a coherent low-dimensional structure. Classical approaches often rely on feature concatenation or explicit alignment assumptions, which become restrictive under heterogeneous geometries or nonlinear distortions. In this work, we propose two geometry-aware multi-view embedding strategies grounded in Gromov-Wasserstein (GW) optimal transport. The first, termed Mean-GWMDS, aggregates view-specific relational information by averaging distance matrices and applying GW-based multidimensional scaling to obtain a representative embedding. The second strategy, referred to as Multi-GWMDS, adopts a selection-based paradigm in which multiple geometry-consistent candidate embeddings are generated via GW-based alignment and a representative embedding is selected. Experiments on synthetic manifolds and real-world datasets show that the proposed methods effectively preserve intrinsic relational structure across views. These results highlight GW-based approaches as a flexible and principled framework for multi-view representation learning.

\end{abstract}



\vspace{0.5em}




\section{Introduction}
\label{Introduction}

Many real-world datasets are naturally described through multiple representations, or views, arising from different sensing modalities, feature extractors, or measurement processes. Effectively integrating such heterogeneous information remains a central challenge in signal processing and machine learning, as naive fusion strategies often fail to capture the underlying structure shared across views.

Multi-view learning has emerged as a principled framework to address this challenge by jointly exploiting complementary representations of the same set of samples. Classical approaches typically rely on feature concatenation or decision-level fusion \cite{yu2025review}. While effective in homogeneous scenarios, these assumptions become restrictive when views exhibit nonlinear distortions, heterogeneous geometries, or complex relational structures, which are usual in manilfolds.

From a geometric perspective, this challenge is closely related to manifold learning, which aims to uncover low-dimensional structures embedded in high-dimensional observation spaces. 

In particular, optimal transport and its extensions provide powerful tools for comparing and aligning distributions through their intrinsic structure \cite{zhang2024learning,lin2023multi, peyre2019computational, van2024distributional, clark2024generalized}. The Gromov-Wasserstein framework, in particular, enables the comparison of distributions supported on distinct metric spaces \cite{Eufrazio, montesuma2023recent} by matching their internal distance relations, making it especially suitable for heterogeneous and multi-view settings.

In this work, we build upon these ideas and investigate geometry-driven strategies for multi-view representation learning based on Gromov-Wasserstein optimal transport. We introduce two complementary strategies for aggregating information across views: a distance averaging approach, denoted as Mean-GWMDS, and a selection-based strategy, denoted as Multi-GWMDS, which identifies a representative structure through consistency across views.

Through extensive experiments on synthetic manifolds and a large-scale real-world electricity consumption dataset, we show that geometry-aware aggregation consistently improves the preservation of relational structure across views. In particular, the proposed strategies exhibit increased robustness under nonlinear distortions when compared to classical multi-view baselines based on linear correlation or Euclidean assumptions.

Overall, the proposed framework provides a flexible and principled approach for integrating heterogeneous views by explicitly accounting for relational geometry, offering improved robustness, interpretability, and compactness in multi-view representation learning.

The remainder of this paper is organized as follows. Section~\ref{sec1} reviews the background concepts required for this work, including  optimal transport with an emphasis on Gromov-Wasserstein formulations and multi-view learning. Section~\ref{sec:methods} introduces the proposed multi-view GWMDS framework and presents the two aggregation strategies, namely Mean-GWMDS and Multi-GWMDS. Experimental results on synthetic manifolds and real-world datasets are reported in Section~\ref{sec:experiments}, analyzing the impact of different dissimilarity choices and aggregation mechanisms. Finally, Section~\ref{sec:conclusion} concludes the paper and discusses potential directions for future research.

\section{Background}
\label{sec1}

This section reviews the fundamental concepts required to support the proposed methodology. We first introduce optimal transport as a geometry-aware framework for comparing probability distributions, followed by the Gromov-Wasserstein
distance, which extends optimal transport to the comparison of distributions supported on distinct metric spaces. Finally, we review the multi-view learning paradigm, which provides the general framework for integrating multiple relational representations of the same data. Together, these concepts establish the theoretical foundation for the proposed approach.


\subsection{Optimal Transport}
\label{Optimal Transport}

\gls{ot} provides a principled framework for comparing probability distributions by explicitly modeling how mass is transported between discrete supports at minimal cost \cite{peyre2019computational}. In signal processing and machine learning, \gls{ot} has emerged as a powerful tool for distributional comparison due to its ability to incorporate geometric information and to define soft correspondences between samples.

Consider two empirical probability distributions supported on finite sets,
\begin{equation}
    \mu = \frac{1}{n}\sum_{i=1}^{n} \delta_{x_i}, \qquad 
    \nu = \frac{1}{m}\sum_{j=1}^{m} \delta_{y_j},
\end{equation}
where $\{x_i\}_{i=1}^n \subset \mathbb{R}^{d_x}$ and $\{y_j\}_{j=1}^m \subset \mathbb{R}^{d_y}$ denote the supports, and $\delta_x$ denotes the Dirac measure centered at $x$.

Given a cost matrix $C \in \mathbb{R}^{n \times m}$, where each entry $C_{ij}$ represents the cost of transporting a unit mass from $x_i$ to $y_j$, the discrete optimal transport problem is formulated as~\cite{peyre2019computational}
\begin{equation}
\pi^{\star} = \argmin{\pi \in \Pi(\mu,\nu)}  \sum_{i=1}^{n} \sum_{j=1}^{m} \pi_{ij} C_{ij},
\end{equation}
where the set of admissible transport plans is defined by
\begin{equation}
\Pi(\mu,\nu) =
\left\{
\pi \in \mathbb{R}_+^{n \times m}
:
\pi \mathbf{1}_m = a, 
\;\;
\pi^\top \mathbf{1}_n = b
\right\},\label{eq:transport-simplex}
\end{equation}
The transport plan \(\pi\) defines a soft alignment between the supports of the two distributions. Rather than enforcing pointwise correspondences, OT allows each element \(x_i\) to distribute its mass across multiple elements \(y_j\), weighted according to the values of \(\pi_{ij}\). 

When the cost matrix \(C \in \mathbb{R}^{n \times m}\) is induced by a metric, such as
\(
C_{ij} = \|x_i - y_j\|^p, \quad p \geq 1,
\)
the optimal value of the Kantorovich transport problem
\begin{equation}
W_p^p(\mu,\nu)
=
\min{\pi \in \Pi(\mu,\nu)}
\sum_{i=1}^n \sum_{j=1}^m \pi_{ij}C_{ij},
\end{equation}
defines the discrete Wasserstein distance of order \(p\) between the empirical measures. This formulation assumes that the elements \(x_i\) and \(y_j\) lie in a common metric space, so that pairwise comparisons across the two supports are well defined. However, in many practical scenarios, this requirement is restrictive. When the supports \(\{x_i\}\) and \(\{y_j\}\) belong to heterogeneous feature spaces, defining a reliable point-wise cost becomes ill-posed or infeasible. This situation happens, for instance, when embeddings are learned independently, or data represents graphs or distance matrices.

These limitations motivate extensions of optimal transport that replace explicit cross-domain costs with comparisons based on intra-domain relational structures, enabling the alignment of distributions supported on distinct spaces. Such extensions are particularly relevant in multi-view and structured data settings and form the basis of Gromov–Wasserstein-type.

\subsubsection{Gromov–Wasserstein Distance}

We start our discussion of the Gromov-Wasserstein distance by defining the notion of metric measure space~\cite{memoli2011gromov}, which is central to this distance.

\begin{definition}\label{def:mm-space}
    A metric measure space is a triple $(\mathcal{X},d_{\mathcal{X}},\mu)$ where $(\mathcal{X},d_{\mathcal{X}})$ is a compact metric space and $\mu$ is a Borel measure on $\mathcal{X}$.
\end{definition}

Consider two metric measure spaces denoted $(\mathcal{X}, d_\mathcal{X}, \mu)$ and $(\mathcal{Y}, D_\mathcal{Y}, \nu)$. Let $X = \{x_{1},\cdots,x_{n}\}$, $x_{i} \in \mathcal{X}$ and $Y = \{ y_{1}, \cdots, y_{m} \}$, $y_{j} \in \mathcal{Y}$. Denote $D_{X} \in \mathbb{R}^{n\times n}$ and $D_{Y} \in \mathbb{R}^{m \times m}$ the intra-domain distance matrices, i.e., $(D_{X})_{i\ell} = d_{\mathcal{X}}(x_{i},x_{\ell})$. The squared Gromov-Wasserstein distance~\cite{kerdoncuff2021sampled} is defined as
\begin{equation}
\mathrm{GW}_{2}^2\!\left( \mu, \nu \right)
=
\min{\pi \in \Pi(\mu,\nu)}
\sum_{i,\ell=1}^{n} \sum_{j,k=1}^{m}
L\!\left((D_X)_{i\ell}, (D_Y)_{jk}\right)
\pi_{ij} \pi_{\ell k}.
\end{equation}
where \(L(\cdot,\cdot)\) is typically chosen as the squared loss \(L(x,y) = (x-y)^2\), and \(\Pi(\mu,\nu)\) denotes the set of couplings satisfying the marginal constraints (cf. equation~\ref{eq:transport-simplex}). The \gls{gw} does not require the two spaces to share the same ambient dimension, coordinate system, or feature representation \cite{kerdoncuff2021sampled}. 

Since the GW distance only depends on the intra-domain distance matrices $D_{X}$ and $D_{Y}$, we adopt an abuse of notation and denote $\text{GW}_{2}(D_{X}, D_{Y})^{2}$, implicitly assuming that $D_{X}$ and $D_{Y}$ come from the pairwise distances in domains $\mathcal{X}$ and $\mathcal{Y}$.

\subsubsection{Gromov–Wasserstein Multidimensional Scaling}\label{sec:gwmds}



Beyond pairwise comparison, the Gromov-Wasserstein framework can be used to learn a geometric representation that preserves the relational structure of a given metric measure space~\cite{Eufrazio}. In this context, Gromov-Wasserstein multidimensional scaling (GWMDS) is a geometry-aware dimensionality reduction technique, closely related to manifold learning methods, whose goal is to recover a low-dimensional representation that preserves intrinsic relational distances. Given an empirical structure $(\mathcal{X}, D_X, \mu)$, the objective is to estimate a configuration $Y \in \mathbb{R}^{n \times q}$ whose induced pairwise distances $D_Y$ are consistent, in the Gromov-Wasserstein sense, with the original distances $D_X$. Formally, this amounts to solving
\begin{equation}
Y^{\star} = \text{GW-MDS}(D_{X}) = \argmin{Y \in \mathbb{R}^{n \times d}} \; \mathrm{GW}_{2}\!\left( D_{X}, D_{Y} \right)^{2},
\end{equation}
where the dependence of $\mathrm{GW}^2(\mu,\nu)$ on the distance matrices $D_X$ and $D_Y$ is implicit. We use GW-MDS$(D_{X})$ to denote the procedure that receives $D_{X}$ as input, and outputs the embedding vectors $Y^{\star}$~\cite[Algorithm 1]{Eufrazio}.



\subsection{Multi-View Learning}
\label{Multi-View}

In the multi-view setting, we are given $v = 1,\cdots, V$ different views of the same objects. We model this problem through the lens of metric measure spaces (cf. Definition~\ref{def:mm-space}): $\{\mathcal{X}^{(v)}, d_{\mathcal{X}^{(v)}}, \mu_{v}\}_{v=1}^{V}$. On each of these spaces, we collect a set of $n$ samples,
\begin{equation}
\mathfrak{X} = \left\{ X^{(1)}, X^{(2)}, \dots, X^{(V)} \right\},
\end{equation}
where each view \( X^{(v)} \in \mathbb{R}^{n \times p_v} \) corresponds to a distinct feature space describing the same underlying samples. Although the dimensionality \( p_v \) and the nature of the features may vary across views, all representations are assumed to be generated from a common phenomenon. The main challenge of this setting is account for the different \emph{geometries} in each measure metric space.


Multi-view learning is motivated by the fact that a single representation often provides only a partial description of complex data. By jointly exploiting multiple views, learning algorithms can leverage complementary information and improve robustness and generalization, particularly in heterogeneous settings
\cite{yu2025review}.

Despite their diversity, most multi-view learning methods rely on two fundamental principles \cite{yu2025review, qin2025survey}. The consensus principle assumes that different views share a common latent structure. In practice, this principle implies that there exists some form of agreement among representations, often modeled by encouraging similarity between view-specific latent variables. Conversely, the complementarity principle acknowledges that each view may encode unique information that cannot be recovered from others. Effective multi-view learning must therefore balance agreement across views while preserving view-specific characteristics, rather than enforcing strict equivalence between representations.

From a methodological perspective, classical multi-view approaches are often distinguished by how information from different views is combined \cite{chowdhury2025deep}. Early fusion strategies aggregate views at the input or representation level, for example by concatenating features into a single vector,
\begin{equation}
x_i = \left[ x_i^{(1)}, x_i^{(2)}, \dots, x_i^{(V)} \right],
\end{equation}
and subsequently applying a standard learning algorithm. While simple, such strategies are sensitive to scale differences, high dimensionality, and redundancy across views. Late fusion strategies instead learn separate models for each view and combine their outputs at the decision level, typically through averaging or voting schemes. Although more flexible, these approaches may fail to exploit cross-view relationships encoded in the data geometry.

To overcome these limitations, a large body of work has focused on multi-view representation learning, where each view \( X^{(v)} \) is mapped to a latent representation \( Z^{(v)} \in \mathbb{R}^{n \times q} \) \cite{chowdhury2025deep, salter2017latent}. The goal is to learn representations that capture shared information across views while maintaining discriminative and compact structures. However, many existing methods implicitly assume that the latent spaces \( Z^{(v)} \) can be aligned through linear transformations or that they lie in compatible Euclidean spaces. Such assumptions become restrictive when views exhibit nonlinear distortions, different intrinsic geometries, or complex relational dependencies \cite{ chowdhury2025deep, xu2013survey}.

In summary, multi-view learning provides a principled framework for integrating heterogeneous representations by exploiting both consensus and complementarity across views. Nevertheless, traditional fusion and representation-level representation-level methods face intrinsic limitations when views are nonlinear, heterogeneous, or structurally complex. Addressing these challenges requires alignment strategies that go beyond feature correspondence and explicitly account for the underlying relational structure of the data, which motivates the geometry-aware approaches explored in the subsequent sections.

\section{Proposed Methods}
\label{sec:methods}

In this section, we introduce two complementary algorithms for the multi-view dimensionality reduction problem.
Both approaches rely on distance-based representations and optimization principles, but differ in how information from multiple views is combined prior to embedding.

The first approach, referred to as Mean-GWMDS, constructs a representative structure by directly combining the distance matrices obtained from the different views.
The second approach, denoted as Multi-GWMDS, follows a selection-based strategy, where a representative embedding is chosen from a set of candidate structures derived from the individual views.

While Mean-GWMDS provides a simple and computationally efficient baseline based on explicit aggregation, Multi-GWMDS offers a more geometry-aware alternative by selecting a structure that best reflects the shared relational patterns across views.
Together, these two methods enable a systematic comparison between averaging-based and selection-based strategies in multi-view geometric embedding.

\subsection{Mean-GWMDS: Distance Averaging Strategy}

The Mean-GWMDS strategy constructs a representative geometric structure by directly aggregating the distance information provided by multiple views. The algorithm can be consulted at \ref{alg:mean_gwmds}. Let
\(
\left\{ D_{X^{(v)}} \in \mathbb{R}^{n \times n} \right\}_{v=1}^{V}
\)
denote the pairwise distance matrices computed independently from each view. These matrices encode the intrinsic geometry of the data within each representation.

In the Mean-GWMDS approach, a single representative distance matrix \(\bar{D}\) is obtained by averaging the view-specific distances:
\begin{equation}
    \bar{D} = \frac{1}{V} \sum_{v=1}^{V} D_{X^{(v)}}.
\end{equation}
This operation is possible, due the special structure of the multi-view problem. For instance, the $i-$th sample from each view, $x_{i}^{(v)}$, corresponds to the same underlying object. Once the averaged distance matrix \(\bar{D}\) is obtained, we use it as input to the \gls{gw-mds} algorithm~\cite[Algorithm 1]{Eufrazio}, described in Section~\ref{sec:gwmds}. Specifically, GW-MDS seeks a low-dimensional embedding $Y \in \mathbb{R}^{n \times q}$ whose pairwise Euclidean distances approximate \(\bar{D}\) in the Gromov-Wasserstein sense. The resulting embedding can be interpreted as a global geometric representation that reflects the mean behavior of the multiple views. This strategy is shown in Algorithm~\ref{alg:mean_gwmds}.

\begin{algorithm}[ht]
\caption{Mean-GWMDS: Distance Averaging Strategy}
\textbf{Input:} Multi-view data $\mathfrak{X}=\{X^{(1)},\ldots,X^{(V)}\}$, embedding dimension $q$, learning rate $\eta$\\
\textbf{Result:} Representation embeddings $Y \in \mathbb{R}^{n \times q}$
\begin{algorithmic}[1]
    \For{$v=1,\ldots,V$}
        \State $D_{X^{(v)}} = \{ d_{\mathcal{X}}(x_{i}^{(v)},x_{j}^{(v)}) \}_{i,j=1}^{n}$
    \EndFor
    \State $\bar{D} \leftarrow \frac{1}{V}\sum_{v=1}^{V} D_{X^{(v)}}$
    \State $Y^{\star} = \text{GW-MDS}(\bar{D})$
    \State \Return $Y^{\star}$
\end{algorithmic}
\label{alg:mean_gwmds}
\end{algorithm}

The main advantage of the Mean-GWMDS strategy lies in its simplicity and computational efficiency. By aggregating distances prior to embedding, the method avoids the need for multiple GW optimizations and provides a natural baseline for multi-view geometric integration. However, this approach may be sensitive to views that exhibit strong distortions or highly heterogeneous geometries, as such effects are directly propagated into the averaged distance matrix.

\subsection{Multi-GWMDS: Selection-Based Strategy}

The Multi-GWMDS strategy adopts a selection-based paradigm in which several geometry-consistent candidate embeddings are first generated, and a representative embedding is subsequently selected based on cross-view relational agreement. This strategy is inspired by the notion of Gromov-Wasserstein barycenters, which aim to summarize multiple relational structures through a common geometric representation~\cite{peyre2016gromov}. Algorithm~\ref{alg:multi_gwmds} summarizes the procedure.

Given a collection of view-specific relational structures $\{D^{(v)}\}_{v=1}^{V}$ and view weights $\lambda \in \Delta_V$, the Multi-GWMDS strategy starts by computing $V$ transport plans and a shared embedding,
\begin{align}
    (Y^{\star}, \{\pi^{(v)}\}_{v=1}^{V}) &= \argmin{ \substack{Y \in \mathbb{R}^{n \times d},\\ \{ \pi^{(v)}\in\Pi(\mu^{(v)},\nu) \}_{v=1}^{V}} }
\sum_{v=1}^{V}\lambda_{v} \sum_{i,k=1}^{n}\sum_{j,\ell=1}^{n}
L(D_{X^{(v)},i,k},D_{Y,j,\ell})
\pi^{(v)}_{ij}\,\pi^{(v)}_{k\ell},\label{eq:joint-embeddings}
\end{align}
where $D_{X^{(v)},i,k}$ denotes $d_{\mathcal{X}}(x_{i}^{(v)},x_{k}^{(v)})$, i.e., the distance between the $i-$th and $k-$th samples in the $\mathcal{X}$. Intuitively, each $\pi^{(v)}_{ij}$ encodes the soft-correspondence between the $j-$th embedding vector $y_{j} \in \mathbb{R}^{q}$ and the $i-$th point of the $v-$th view, $x_{i}^{(v)}$. Equation~\ref{eq:joint-embeddings} implies that embeddings might correspond to different points in the views. We then align the views via,
\begin{align}
    y^{(v)}_{k} = \sum_{i=1}^{n}\dfrac{\pi_{ik}^{(v)}}{\sum_{j=1}^{n}\pi_{jk}^{(v)}}y_{i}\text{, and }v^{\star} = \argmin{v=1,\cdots,V}\dfrac{1}{V}\sum_{u=1}^{V}\rho(D_{X^{(u)}},D_{Y^{(v)}}),\label{eq:alignment-choosing}
\end{align}
in other words, we choose the alignment that results in the most plausible embeddings, on average. Equation~\ref{eq:alignment-choosing} can be interpreted as the multi-view analogue of the embedding alignment of~\cite[Algorithm 1, Line 8]{Eufrazio}. The overall strategy is shown in Algorithm~\ref{alg:multi_gwmds}.



\begin{algorithm}[ht]
\caption{Multi-GWMDS: Selection-Based Strategy}
\textbf{Input:} Multi-view data $\mathfrak{X}=\{X^{(1)},\ldots,X^{(V)}\}$, embedding dimension $d$, learning rate $\eta$\\
\textbf{Result:} Selected embedding $Y^{(v^\star)} \in \mathbb{R}^{n \times d}$
\begin{algorithmic}[1]
    \For{$v=1,\ldots,V$}
        \State $D^{(v)} \leftarrow d_X(X^{(v)})$
    \EndFor
    \State $(Y^{\star}, \{ \pi^{(v)} \}_{v=1}^{V})$ minimizing Equation~\ref{eq:joint-embeddings}
    \State Create view-aligned embeddings $Y^{(v)}$ using equation~\ref{eq:alignment-choosing}
    \State $v^{\star} \leftarrow \argmin{v=1,\cdots,V}\dfrac{1}{V}\sum_{u=1}^{V}\rho(D_{X^{(u)}}, D_{Y^{(v)}})$
    \State \Return $Y^{(v^{\star})}$
\end{algorithmic}
\label{alg:multi_gwmds}
\end{algorithm}


The two proposed strategies are designed to address complementary multi-view scenarios. The Mean-GWMDS approach is particularly effective when the views exhibit comparable geometric structures and similar levels of distortion, as direct averaging at the distance level provides a stable and computationally efficient representation of the shared geometry. In contrast, the Multi-GWMDS approach is tailored for heterogeneous settings, where views may differ significantly in terms of noise, deformation, or underlying relational structure. By selecting a representative structure based on a robust aggregation of view-wise correlations, Multi-GWMDS mitigates the influence of atypical or degraded views and favors a consensus geometry that is maximally consistent across the ensemble.


\section{Experiments and Discussion}\label{sec:experiments}

This section presents an experimental evaluation of the proposed methods on both synthetic and real-world datasets. The experiments are designed to assess the ability of the proposed strategies to capture and preserve relational structures across multiple views. 

In particular, we consider the Electricity Load Diagrams (ELD) dataset \cite{electricityloaddiagrams20112014_321}, publicly available from the UCI (University of California, Irvine) Machine Learning Repository \cite{kelly2023uci}. In dataset is particularly well suited for evaluating robustness to heterogeneity, as consumption profiles may vary significantly across time periods due to seasonal effects, external factors, and user-specific behavior \cite{shi2020approach}.

\subsection{Manifold experiments}

Within this context, synthetic manifold datasets play a central role in evaluating representation learning methods. By considering classical nonlinear manifolds with known intrinsic geometry, such as those commonly used in the literature (Swiss Roll and S-curve), it becomes possible to systematically assess how well a given method preserves geometric relationships under controlled distortions. These datasets therefore provide an ideal testbed for analyzing the geometric behavior of the proposed strategies before moving to real-world scenarios.

We consider several synthetic manifolds, including the S-curve, the Swiss roll, and the Mobius strip. These datasets exhibit distinct geometric and topological properties, ranging from smooth nonlinear embeddings in Euclidean space to non-orientable surfaces.

For each manifold, view-specific relational structures are constructed and used as inputs to the proposed multi-view strategies. Both Mean-GWMDS and Multi-GWMDS are applied to integrate information across views, and the resulting embeddings are analyzed in terms of structure preservation and robustness to view heterogeneity.

In the manifold experiments, a two-view setting is considered to analyze the behavior of the proposed strategies under controlled geometric transformations. Given an original manifold, two distinct views are constructed:

The first view is produced by applying a rigid rotation of $40^\circ$ around the $z$-axis.
Let $\theta = 40^\circ$ and define the rotation matrix
\begin{equation}
X^{(1)} = X\,R(\theta)^{\top}\text{, where }R(\theta)=
\begin{pmatrix}
\cos\theta & -\sin\theta & 0 \\
\sin\theta & \cos\theta  & 0 \\
0          & 0           & 1
\end{pmatrix}.
\end{equation}
The second view is generated through a linear deformation combining scaling (elongation) and shear 
\begin{equation}
X^{(2)} = X\,A^{\top}\text{, where }A =
\begin{pmatrix}
1.8 & 0.3 & 0 \\
0   & 1.0 & 0 \\
0   & 0   & 0.6
\end{pmatrix}
\end{equation}
These transformations preserve the underlying sample correspondence while inducing heterogeneous geometric distortions across views.

For each view, pairwise dissimilarities are initially computed using the Euclidean distance, providing a baseline relational representation, as reported in Table~\ref{tab:correlation_multiview_euclidian}.  The proposed methods are compared against several classical multi-view baselines, including Canonical Correlation Analysis (CCA) \cite{hotelling1992relations, sun2013survey}, Multiset Canonical Correlation Analysis (MCCA) \cite{kettenring1971canonical}, and multi-view multidimensional scaling (MVMDS) \cite{bai2017multidimensional}.

As shown in Table~\ref{tab:correlation_multiview_euclidian}, Mean-GWMDS achieves the highest average correlations across most synthetic manifolds, indicating that averaging view-specific relational structures provides a strong and stable multi-view representation when Euclidean distances are employed. Multi-GWMDS also attains competitive performance, particularly on individual views, while offering a more flexible alignment strategy across views. In contrast, Euclidean and correlation-based baselines may perform well on specific views but exhibit less consistent behavior when aggregating relational information across views.

\begin{table}[h]
    \centering
    \begin{tabular}{lcccccc}
        \toprule
        & & \shortstack{Multi\\GWMDS} & \shortstack{Mean\\GWMDS} & CCA & MCCA & MVMDS \\
        \midrule
       S-curve & view 1 & \textbf{0.8484} & 0.7750 & 0.7776 & 0.7726 & 0.7748 \\
               & view 2 & 0.8530 & \textbf{0.9334} & 0.8878 & 0.8545 & 0.8491 \\
               & mean   & 0.8507 & \textbf{0.8542} & 0.8327 & 0.8135 & 0.8119 \\
    
        \midrule
       Swiss Roll & view 1 & 0.7762 & 0.8095 & 0.7620 & 0.7418 & \textbf{0.8235} \\
                  & view 2 & \textbf{0.9289} & 0.9055 & 0.8066 & 0.4527 & 0.8872 \\
                  & mean   & 0.8525 & \textbf{0.8575} & 0.7843 & 0.5973 & 0.8554 \\

        \midrule
       Mobius & view 1 & 0.9205 & 0.9231 & \textbf{0.9293} & 0.7106 & 0.9292 \\
              & view 2 & \textbf{0.9660} & 0.9644 & 0.8931 & 0.6591 & 0.8830 \\
              & mean   & 0.9433 & \textbf{0.9437} & 0.9112 & 0.6849 & 0.9061 \\

        \midrule
       Torus  & view 1 & 0.6481 & 0.7970 & 0.6671 & \textbf{0.8304} & 0.8257 \\
              & view 2 & \textbf{0.9431} & 0.8284 & 0.8622 & 0.3726 & 0.7840 \\
              & mean   & 0.7956 & \textbf{0.8127} & 0.7647 & 0.6015 & 0.8049 \\
        \bottomrule
    \end{tabular}
    \caption{Correlation results for each view across different multi-view methods. For Multi-GWMDS and Mean-GWMDS, Euclidean distances are used to construct the relational structures in each view.}
    \label{tab:correlation_multiview_euclidian}
\end{table}

\begin{figure}[H]
    \centering
    \begin{subfigure}[t]{0.45\textwidth}
        \centering
        \includegraphics[width=\textwidth]{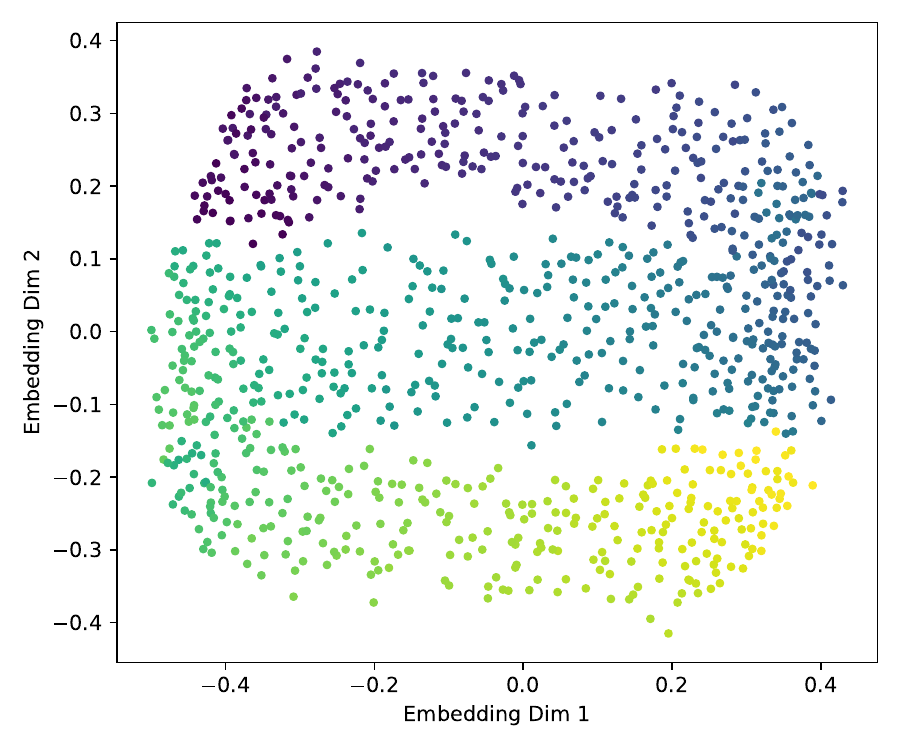}
        \caption{Multi-GWMDS}
        \label{fig:scurve_multigwmds}
    \end{subfigure}
    \hfill
    \begin{subfigure}[t]{0.45\textwidth}
        \centering
        \includegraphics[width=\textwidth]{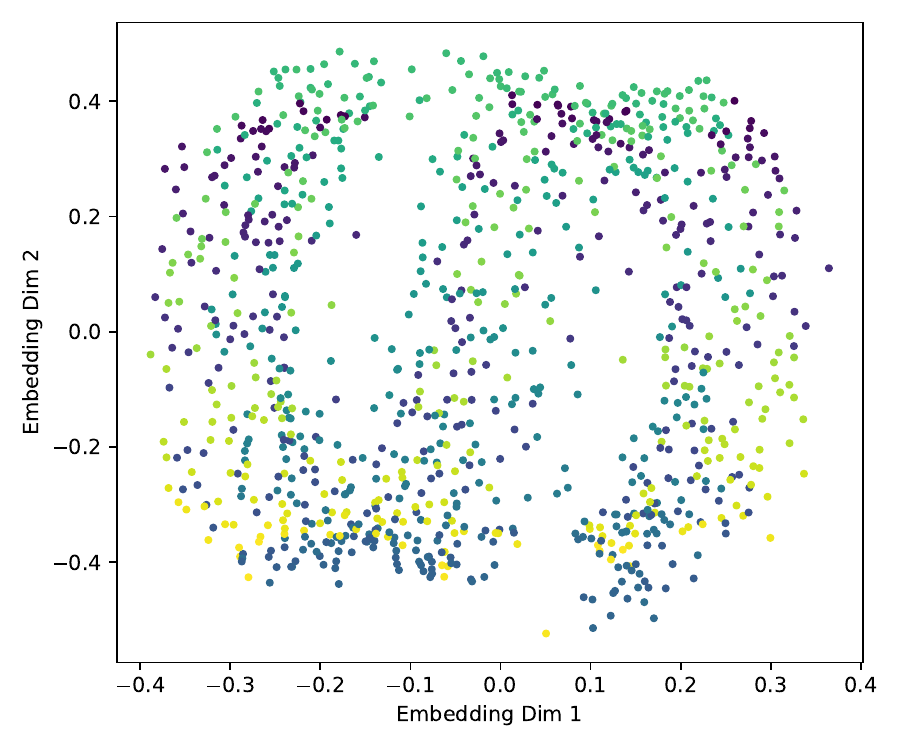}
        \caption{Mean-GWMDS}
        \label{fig:scurve_meangwmds}
    \end{subfigure}

    \vspace{0.3cm}

    \begin{subfigure}[t]{0.32\textwidth}
        \centering
        \includegraphics[width=\textwidth]{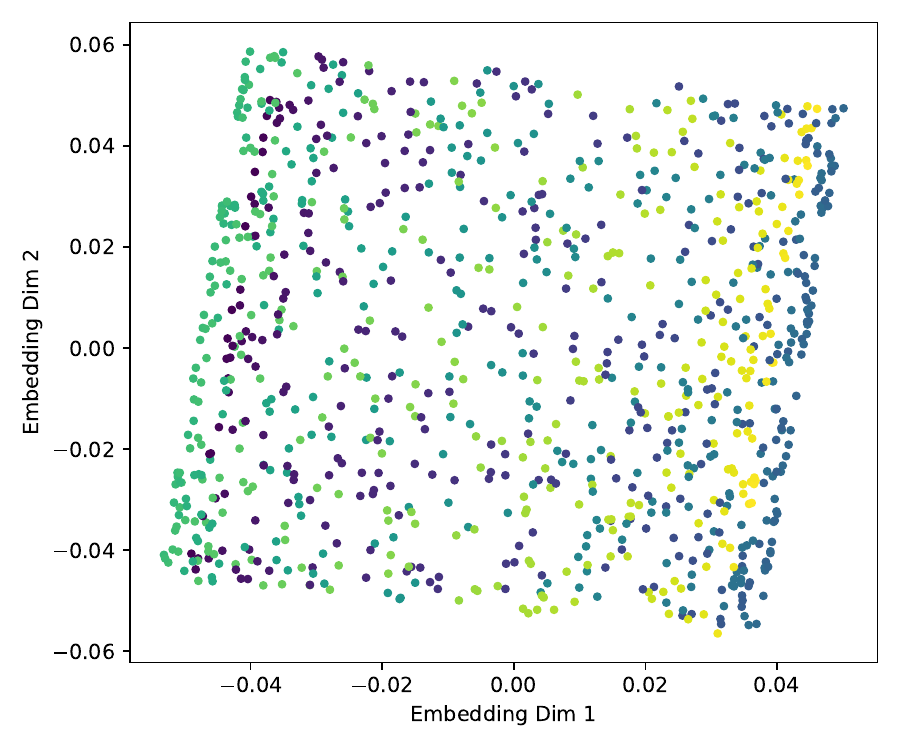}
        \caption{MVMDS}
        \label{fig:scurve_mvmds}
    \end{subfigure}
    \hfill
    \begin{subfigure}[t]{0.32\textwidth}
        \centering
        \includegraphics[width=\textwidth]{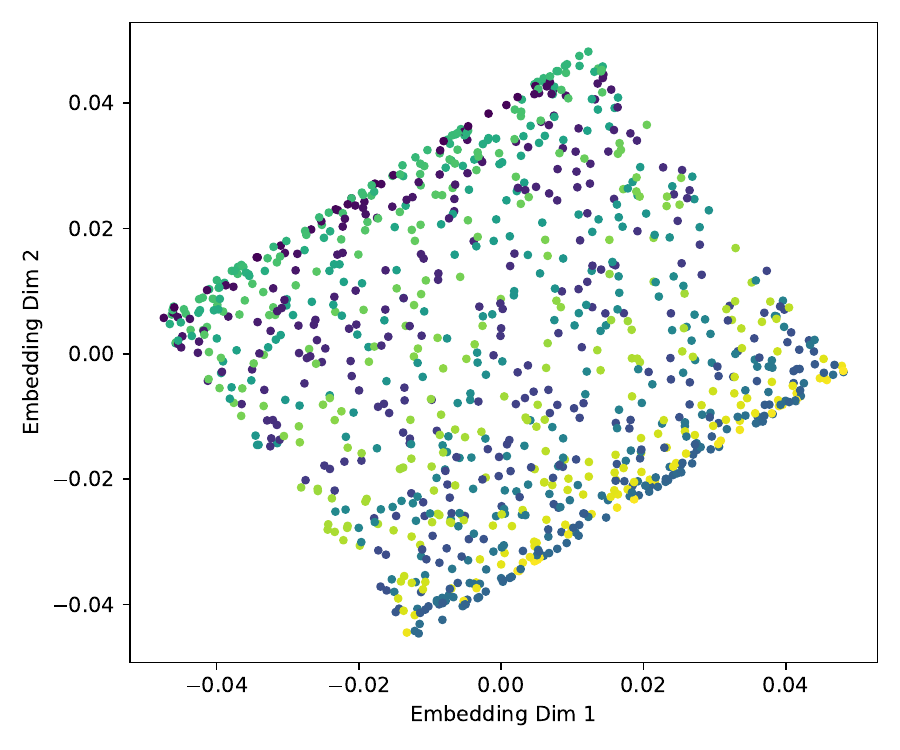}
        \caption{MCCA}
        \label{fig:scurve_mcca}
    \end{subfigure}
    \begin{subfigure}[t]{0.32\textwidth}
        \centering
        \includegraphics[width=\textwidth]{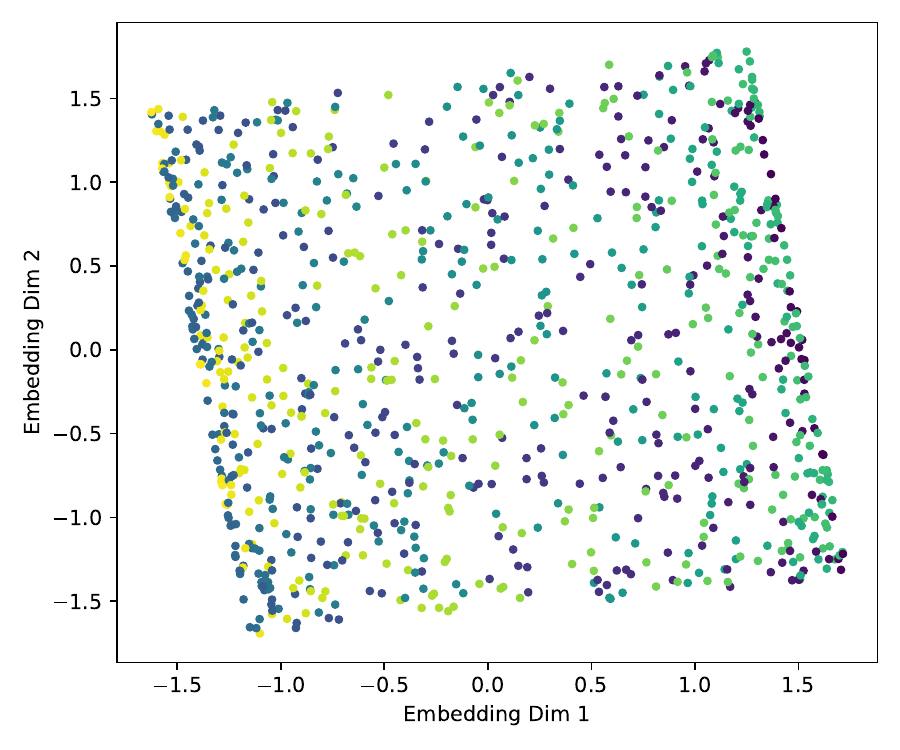}
        \caption{CCA}
        \label{fig:scurve_cca}
    \end{subfigure}


    \caption{Embeddings obtained on the S-curve dataset using different multi-view methods.}
    \label{fig:scurve_2x2}
\end{figure}

Figure~\ref{fig:scurve_2x2} presents a qualitative comparison of embeddings obtained on the S-curve dataset using different multi-view methods under Euclidean dissimilarities. The Multi-GWMDS embedding (top-left) exhibits a clear preservation of the characteristic nonlinear structure of the S-curve, indicating that the proposed framework is able to integrate information across views while maintaining global geometric coherence. The Mean-GWMDS embedding (top-right) also captures the overall manifold structure and produces a visually coherent representation, suggesting that averaging view-specific relational information already yields a stable and faithful embedding. However, subtle distortions and local irregularities can be observed when compared to Multi-GWMDS, reflecting the limitations of simple averaging in the presence of view-dependent geometric variations.

The MVMDS embedding, shows noticeable distortions, particularly in regions where nonlinear geometry plays a dominant role. This behavior reflects the limitations of distance averaging and stress-based optimization when operating purely under Euclidean assumptions in multi-view settings. The embeddings produced by MCCA and CCA further highlight the challenges faced by correlation-based approaches. While these methods preserve certain ordering or variance directions in the data, they fail to recover the intrinsic curved geometry of the manifold, resulting in representations that do not faithfully reflect the underlying structure of the S-curve. 

Overall, this comparison illustrates the advantage of explicitly modeling relational alignments across views. Even when restricted to Euclidean distances, Multi-GWMDS benefits from view-dependent transport couplings, leading to an improved preservation of nonlinear manifold structure when compared to classical multi-view baselines.
    




Next, we extend the manifold learning experiments by replacing Euclidean dissimilarities with geodesic distances in order to better capture the intrinsic geometry of the data. For each view, we first construct a $k$-nearest neighbor graph using the original representation, where each node corresponds to a sample and edges connect pairs of samples that are among the $k$ closest according to the Euclidean distance.

Once the neighborhood graph is built, geodesic distances between all pairs of samples are approximated by computing the shortest path distances on this graph. This procedure effectively estimates the length of the paths that lie along the underlying data manifold, rather than relying on direct Euclidean distances that may cut through low-density regions. The resulting shortest path distances define a view-specific geodesic distance matrix, which is subsequently used.

In this setting, the proposed strategies are evaluated using geodesic distance matrices computed independently for each view.

Both variants of the proposed framework, namely Mean-GWMDS and Multi-GWMDS, are applied under this geodesic formulation. This allows us to assess their ability to integrate nonlinear geometric information across views while preserving the intrinsic manifold structure. For comparison, we consider the Multi-Isomap \cite{rodosthenous2024multi} approach as a reference baseline, see Table \ref{tab:correlation_multiview_geodesica}. Multi-Isomap  extends the classical Isomap algorithm to the multi-view setting by aggregating neighborhood graphs or geodesic distances across views prior to embedding. As such, it provides a natural benchmark for evaluating geometry-based multi-view methods that rely on geodesic information.

\begin{figure}[H]
    \centering
    \begin{subfigure}[t]{0.32\textwidth}
        \centering
        \includegraphics[width=\textwidth]{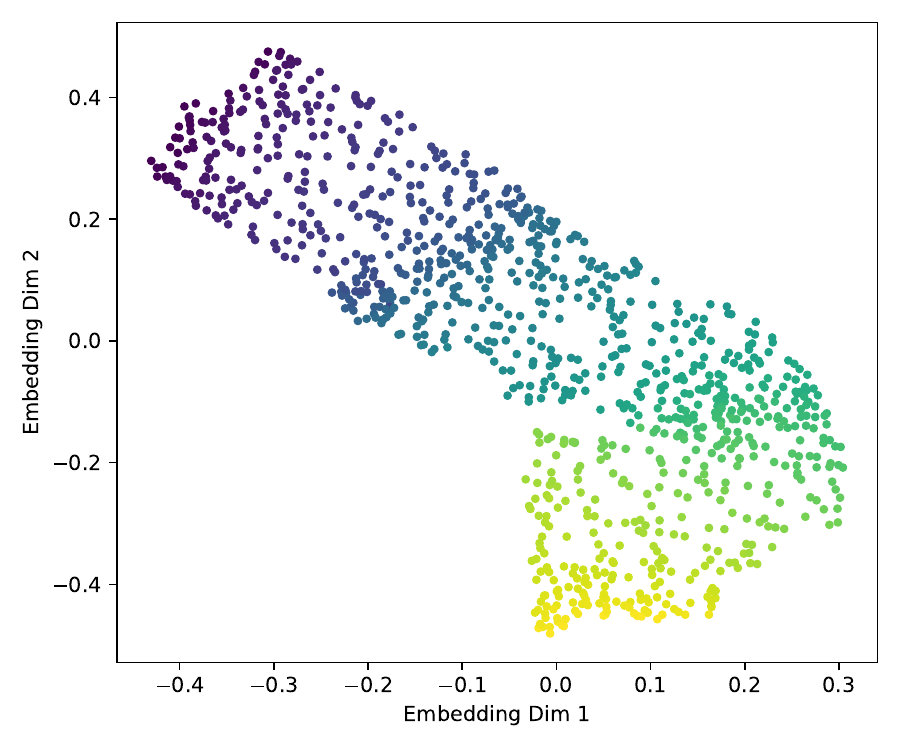}
        \caption{Multi-GWMDS}
        \label{fig:scurve_multigwmds2}
    \end{subfigure}
    \hfill
    \begin{subfigure}[t]{0.32\textwidth}
        \centering
        \includegraphics[width=\textwidth]{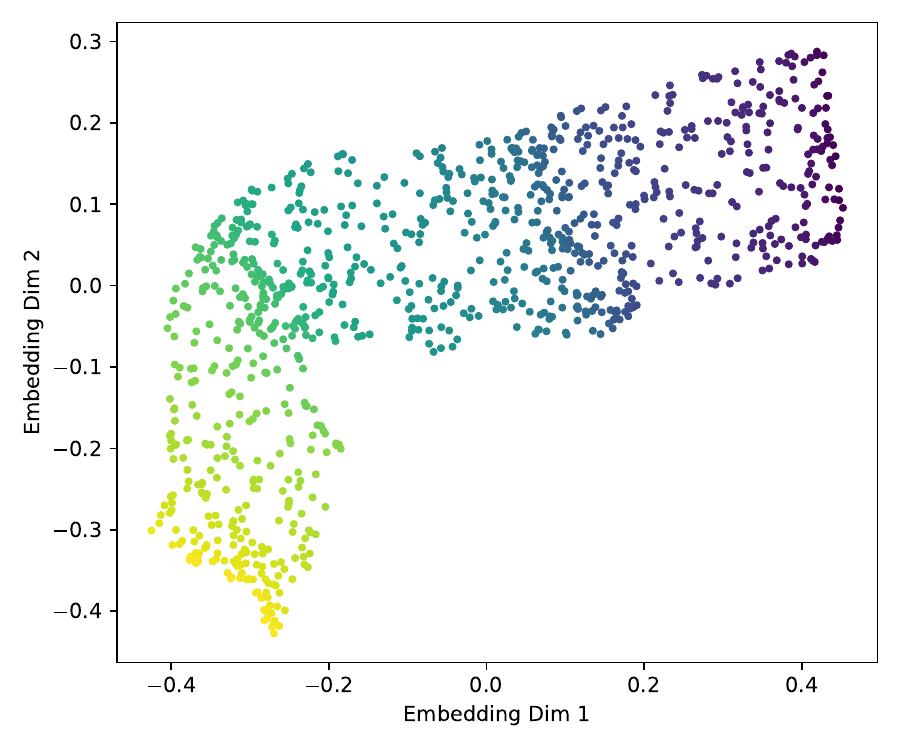}
        \caption{Mean-GWMDS}
        \label{fig:scurve_meangwmds2}
    \end{subfigure}
    \hfill
    \begin{subfigure}[t]{0.32\textwidth}
        \centering
        \includegraphics[width=\textwidth]{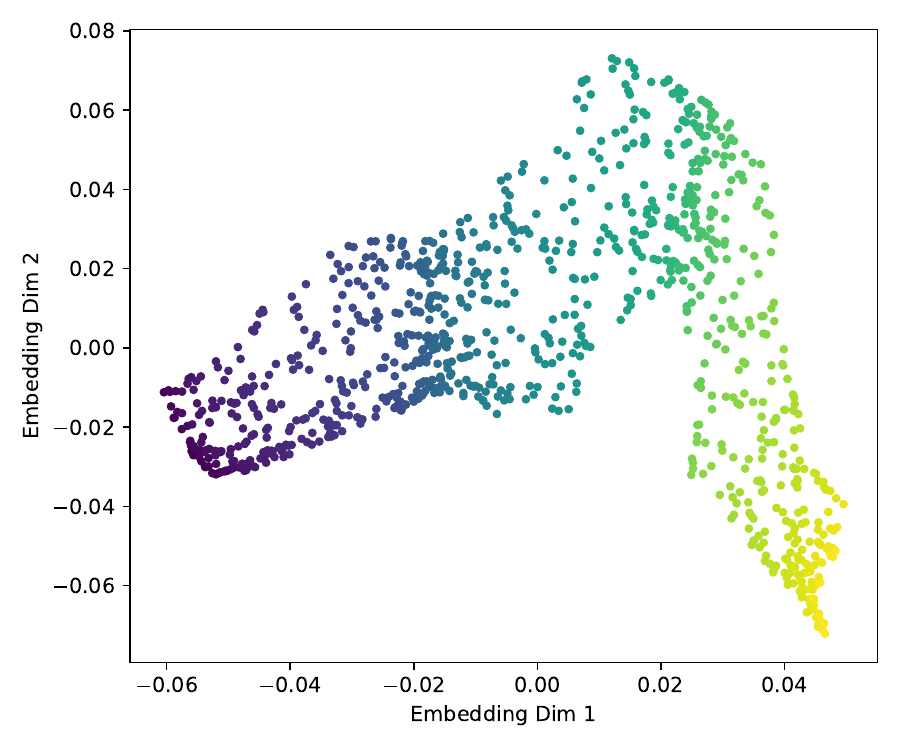}
        \caption{Multi-Isomap}
        \label{fig:scurve_multiisomap}
    \end{subfigure}

    \caption{Embeddings obtained on the S-curve dataset using different multi-view strategies with geodesic distance.}
    \label{fig:scurve_3methods}
\end{figure}

By adopting a common geodesic distance construction across all methods, this experimental setup ensures a fair comparison focused on the ability of each approach to recover the underlying manifold geometry under. Figure~\ref{fig:scurve_3methods} presents a comparison of embeddings obtained on the S-curve dataset using geodesic.

\begin{table}[h]
    \centering
    \begin{tabular}{lccccc}
        \toprule
        & & Multi-GWMDS & Mean-GWMDS & Multi-Isomap \\
        \midrule
       S-curve & view 1 & 0.9556 & \textbf{0.9788} & 0.8009  \\
               & view 2 & \textbf{0.9811} & 0.9765 & 0.8197 \\
               & mean & 0.9684 & \textbf{0.9777} & 0.8103 &  \\
        \midrule
       Swiss Roll & view 1 & \textbf{0.9771} & 0.8507 & 0.7472 \\
                  & view 2 & 0.3002 & 0.6452 & \textbf{0.6833} \\
                  & mean & 0.6387 & \textbf{0.7480} & 0.7153  \\
        \midrule
       Mobius & view 1 & 0.9406 & \textbf{0.9412} & 0.9386 \\
              & view 2 & \textbf{0.9714} & 0.9711 & 0.9568 \\
              & mean& 0.9560 & \textbf{0.9562} & 0.9477  \\
        \midrule
       Torus  & view 1 & 0.7430 & \textbf{0.7546}  &  0.7500 \\
              & view 2 & 0.9261 & \textbf{0.9348}  &  0.9068 \\
              & mean & 0.8346 & \textbf{0.8447} &  0.8284 \\
        \bottomrule
    \end{tabular}
    \caption{Correlation results for each view across different multi-view methods. For all methods, relational structures are constructed from geodesic distance.}
    \label{tab:correlation_multiview_geodesica}
\end{table}

Overall, this comparison highlights that while geodesic distances substantially improve manifold recovery for all methods, explicitly modeling and selecting among view-dependent relational alignments, as done in Multi-GWMDS, provides a more faithful and robust representation of the underlying geometry.

\subsection{Real-World Experiments: Electricity Load Diagrams}

We use the Electricity Load Diagrams 2011–2014 dataset from the UCI Machine Learning Repository, which contains electricity consumption measurements collected over a four-year period. The dataset comprises 370 individual customers, each represented by a high-resolution electricity load time series recorded at 15-minute intervals between 2011 and 2014.

After removing the timestamp information, the full dataset can be represented as a matrix, $X \in \mathbb{R}^{370 \times 140256}$, where each row corresponds to a customer and each column corresponds to a time step. This representation highlights the extreme dimensionality and strong temporal structure of the data.

To construct a multi-view setting, we consider a daily decomposition of the time series. Specifically, four distinct days are selected, and each day is treated as an individual view of the data. Each daily view can thus be represented as a matrix, $X^{(v)} \in \mathbb{R}^{370 \times 96}$, where rows correspond to customers and columns correspond to the 15-minute time slots within a 24-hour period. This formulation allows each day to be interpreted as a multivariate snapshot capturing the joint electricity consumption patterns of all customers over a single day. By treating different days as distinct but related views, the dataset naturally induces a multi-view structure with shared instances and view-dependent temporal characteristics.

We next evaluate the proposed framework on the Electricity Load Diagrams dataset using geodesic dissimilarities in order to capture the intrinsic temporal structure of electricity consumption patterns.

For each view, a neighborhood graph is built and geodesic distances are computed, yielding a view-specific relational structure that reflects nonlinear temporal dependencies. This experimental setup enables the assessment of the proposed strategies in a realistic and large-scale scenario, where temporal heterogeneity and nonlinear structure are naturally present in the data.

\begin{figure}[h]
    \centering
    \begin{subfigure}[t]{0.32\linewidth}
        \centering
        \includegraphics[width=\linewidth]{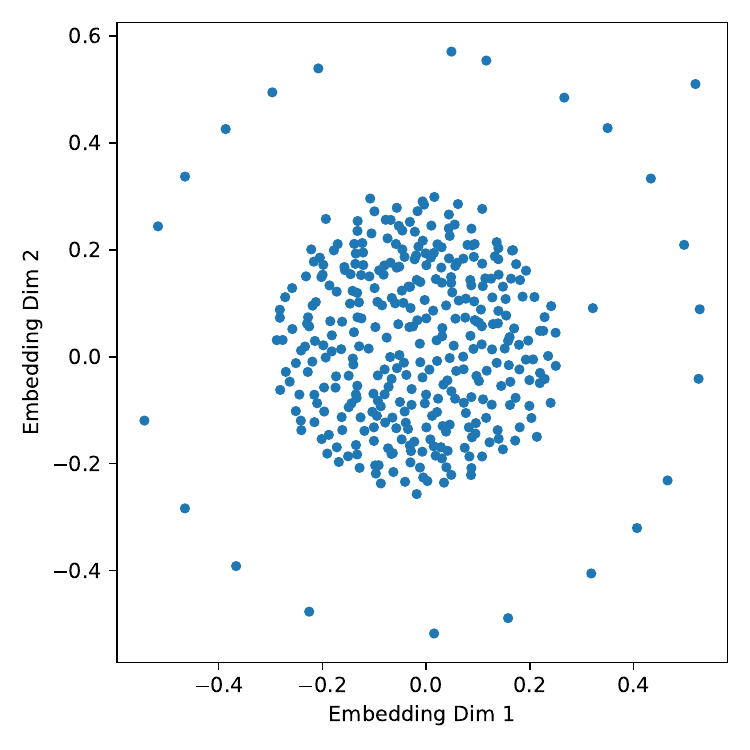}
        \caption{Mean-GWMDS}
        \label{fig:DADOS_REAIS_MEDIO_GWMDS}
    \end{subfigure}
    \hfill
    \begin{subfigure}[t]{0.32\linewidth}
        \centering
        \includegraphics[width=\linewidth]{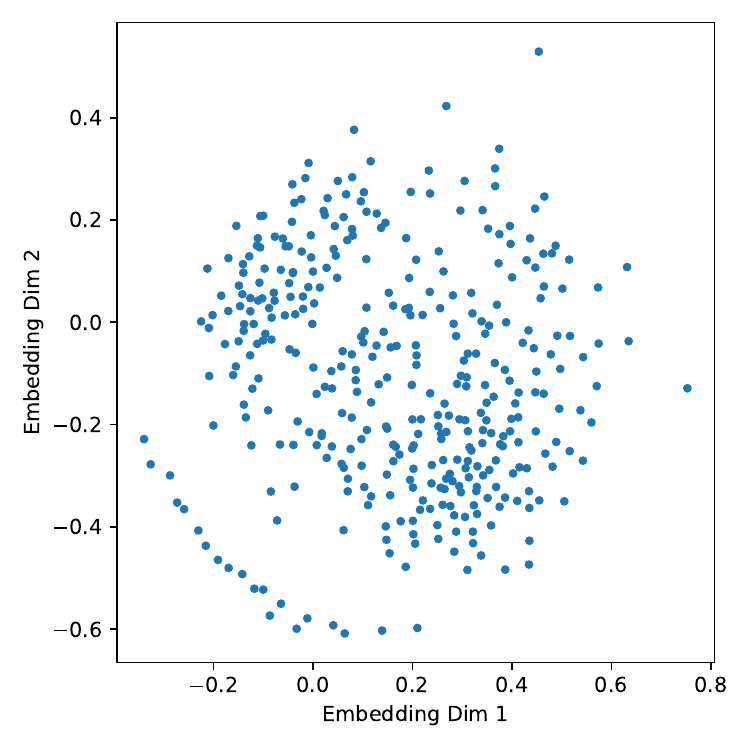}
        \caption{Multi-GWMDS}
        \label{fig:Geodésica/DADOS_REAIS_Multi_GWMDS}
    \end{subfigure}
    \hfill
    \begin{subfigure}[t]{0.32\linewidth}
        \centering
        \includegraphics[width=\linewidth]{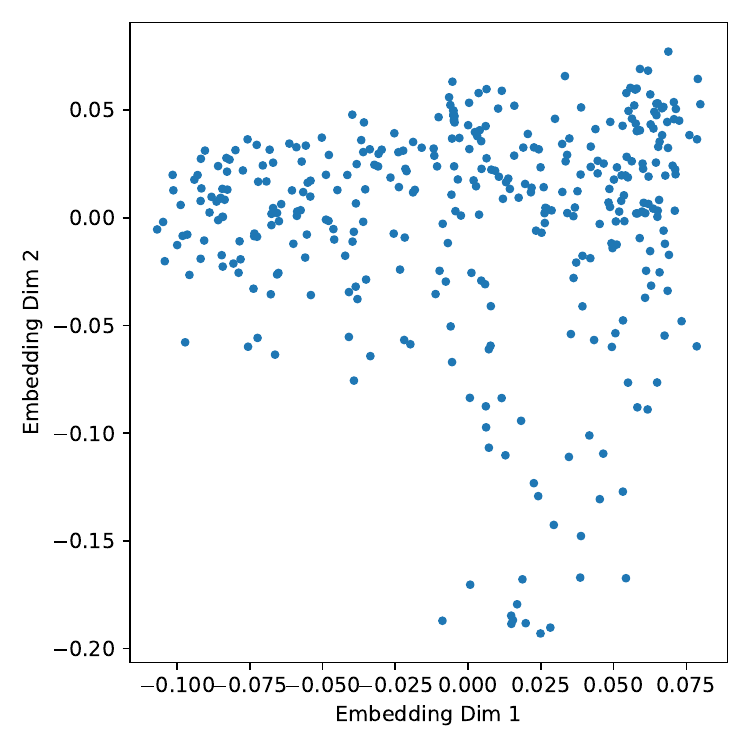}
        \caption{Multi-Isomap}
        \label{fig:real_multi_isomap}
    \end{subfigure}

    \caption{Low-dimensional embeddings of the Electricity Load Diagrams (ELD) dataset obtained using different multi-view methods under geodesic distances.}
    \label{fig:eld_geodesic_comparison}
\end{figure}

Figure~\ref{fig:DADOS_REAIS_MEDIO_GWMDS} shows the embedding obtained using the Mean-GWMDS. Although a single distance matrix is used at this stage, the resulting embedding still reflects the multi-view nature of the data, as the averaged structure encodes shared geometric information across views. The points located farther from the main cluster in the Mean-GWMDS embedding do not necessarily correspond to customers with higher energy consumption. Instead, these points represent customers whose daily consumption profiles exhibit relational patterns that are less similar to those of the majority. 

\begin{figure}
    \centering
    \includegraphics[width=1\linewidth]{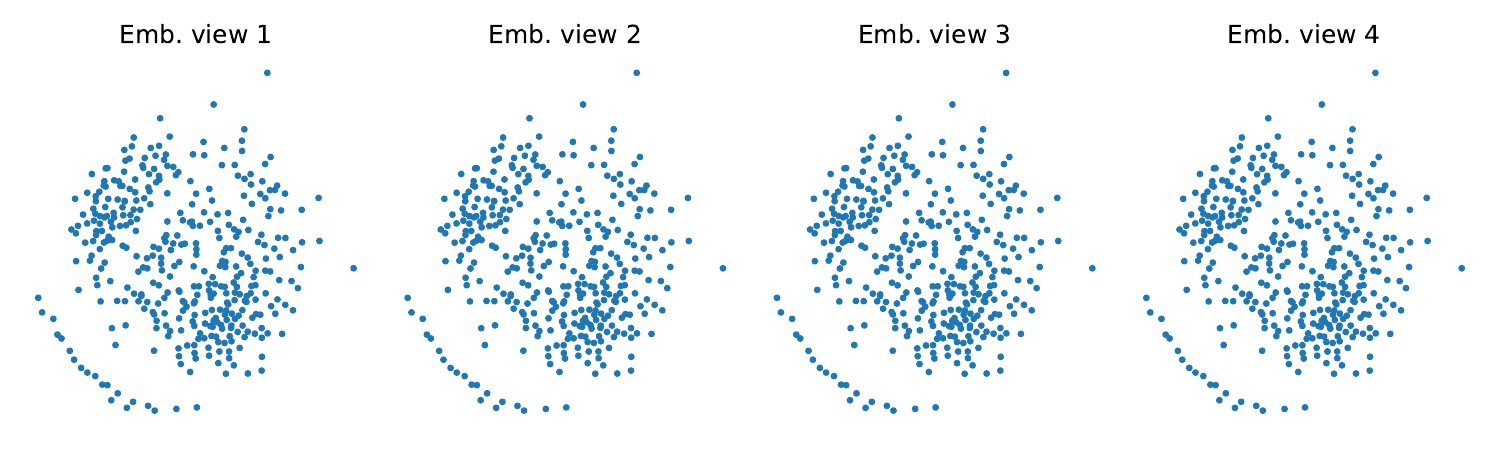}
    \caption{Low-dimensional embeddings of the Electricity Load Diagrams (ELD) dataset obtained with Multi-GWMDS using geodesic distances. Each panel corresponds to an embedding induced by a different view.}
    \label{fig:DADOS REAIS_VARIOS_VIEWS_Multi_GWMDS}
\end{figure}

Figure~\ref{fig:DADOS REAIS_VARIOS_VIEWS_Multi_GWMDS} illustrates a key aspect of the proposed Multi-GWMDS framework. Although all embeddings are learned with respect to the same reference distance structure, each view induces a distinct optimal transport plan between the original space and the representation space. As a consequence, even when the target dimensionality and the distance definition are fixed, the resulting embeddings differ due to the view-specific transport couplings.

This variability reflects the fact that each view emphasizes different relational patterns, which are encoded in its corresponding transport matrix. A representative embedding is then selected based on cross-view consistency, quantified through correlation measures between distance matrices. Depending on the application, this consistency can be aggregated using different criteria, such as the mean, the median, or a max-min strategy, each reflecting a different notion of robustness across views.

This selection mechanism constitutes the core idea of the Multi-GWMDS approach, instead of collapsing view-specific geometries prematurely, it leverages the diversity of transport plans to identify the embedding that best preserves the shared relational structure across views.


Figure~\ref{fig:Geodésica/DADOS_REAIS_Multi_GWMDS} presents the representative embedding obtained with the  Multi-GWMDS. The embedding shown corresponds to the selected representation after evaluating cross-view consistency through correlation analysis. The final embedding is chosen as the one that maximizes this consistency measure, ensuring that the selected representation best preserves the shared relational structure across views.

Table~\ref{tab:eld_correlation_vertical} reports correlation results on the Electricity Load Diagrams dataset using geodesic dissimilarities. The results highlight distinct behaviors among the evaluated multi-view strategies. In contrast to averaging-based strategies, the Multi-GWMDS embedding \ref{fig:Geodésica/DADOS_REAIS_Multi_GWMDS}  exhibits a broader dispersion, reflecting the preservation of view-dependent relational variability across different days. This behavior highlights the ability of Multi-GWMDS to expose heterogeneous temporal structures by explicitly evaluating and selecting among multiple relational alignments.


In Table~\ref{tab:eld_correlation_vertical} the Multi-Isomap baseline consistently yields lower correlations across all views, reflecting the limitations of graph-averaging approaches in capturing heterogeneous temporal or structural variations present in real-world consumption data. While Multi-Isomap (Figure~\ref{fig:real_multi_isomap}) is able to preserve a smooth global geometry through graph averaging, it lacks an explicit mechanism to assess view consistency, leading to oversmoothing and reduced sensitivity to temporal heterogeneity when compared to transport-based multi-view strategies such as Mean-GWMDS and Multi-GWMDS.

Overall, these results confirm the complementary nature of the proposed strategies: Mean-GWMDS favors robustness and global consistency, while Multi-GWMDS excels in retaining distinctive relational structures present in individual views.


\begin{table}[t]
    \centering
    \begin{tabular}{lcccc}
        \toprule
        & \multicolumn{4}{c}{\textbf{Geodesic  distances}} \\
        \cmidrule(lr){2-5}
        & Multi-GWMDS & Mean-GWMDS & Multi-Isomap & \\
        \midrule
        view 1 & 0.3342 & \textbf{0.4814} & 0.2956 &  \\
        view 2 & 0.3891 & \textbf{0.5004} & 0.3484 &  \\
        view 3 & \textbf{0.7913} & 0.4771 & 0.3697 &  \\
        view 4 & 0.3950 & \textbf{0.5036} & 0.3744 &  \\
        mean   & 0.4774 & \textbf{0.4906} & 0.3470 &  \\

        \midrule\midrule

        & \multicolumn{4}{c}{\textbf{Euclidean distances}} \\
        \cmidrule(lr){2-5}
        & Multi-GWMDS & Mean-GWMDS & MVMDS & MCCA \\
        \midrule
        view 1 & 0.2990 & \textbf{0.3921} & 0.3713 & 0.3312 \\
        view 2 & 0.3891 & \textbf{0.4198} & 0.3910 & 0.3399 \\
        view 3 & 0.3836 & \textbf{0.4139} & 0.3767 & 0.3296 \\
        view 4 & \textbf{0.7789} & 0.4384 & 0.4347 & 0.3787 \\
        mean   & \textbf{0.4627} & 0.4161 & 0.3934 & 0.3449 \\
        \bottomrule
    \end{tabular}
    \caption{Correlation between the low-dimensional embeddings and the ground-truth temporal manifold coordinates on the Electricity Load Diagrams (ELD) dataset. Results are reported using geodesic distances (top block) and Euclidean distances (bottom block).}
    \label{tab:eld_correlation_vertical}
\end{table}




Including Euclidean distances allows us to assess the sensitivity of the proposed methods to the choice of relational representation and to contrast topology-aware and feature-based notions of similarity on the same data, see Table \ref{tab:eld_correlation_vertical}. Table reports correlation results on the Electricity Load Diagrams dataset using Euclidean dissimilarities. The results reveal distinct behaviors among the evaluated methods across different views.

Mean-GWMDS achieves the highest correlations on most individual views (views 1, 2, and 3), indicating that averaging view-specific relational structures provides an effective representation when Euclidean distances are employed. Multi-GWMDS, in turn, attains the highest correlation on the most structurally informative view (view 4) and achieves the best average performance across all views. MVMDS yields competitive but consistently lower correlations, reflecting the limitations of directly preserving Euclidean distances in heterogeneous multi-view settings. MCCA exhibits the lowest correlations across all views, which is expected given its reliance on linear correlations in the feature space and the absence of an explicit distance-based objective. Overall, these results suggest that while Euclidean baselines may perform well on individual views, the proposed Multi-GWMDS framework provides superior robustness when aggregating heterogeneous relational information across multiple views.

\begin{figure}[H]
    \centering
    \begin{subfigure}[t]{0.45\textwidth}
        \centering
        \includegraphics[width=\textwidth]{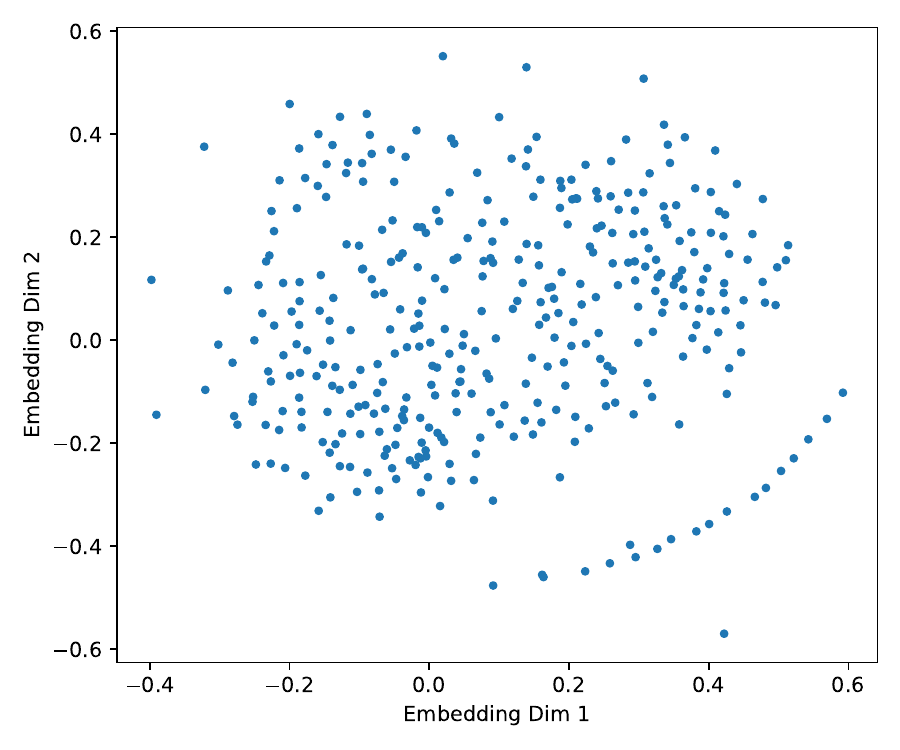}
        \caption{Multi-GWMDS}
        \label{fig:eld_multigwmds}
    \end{subfigure}
    \hfill
    \begin{subfigure}[t]{0.45\textwidth}
        \centering
        \includegraphics[width=\textwidth]{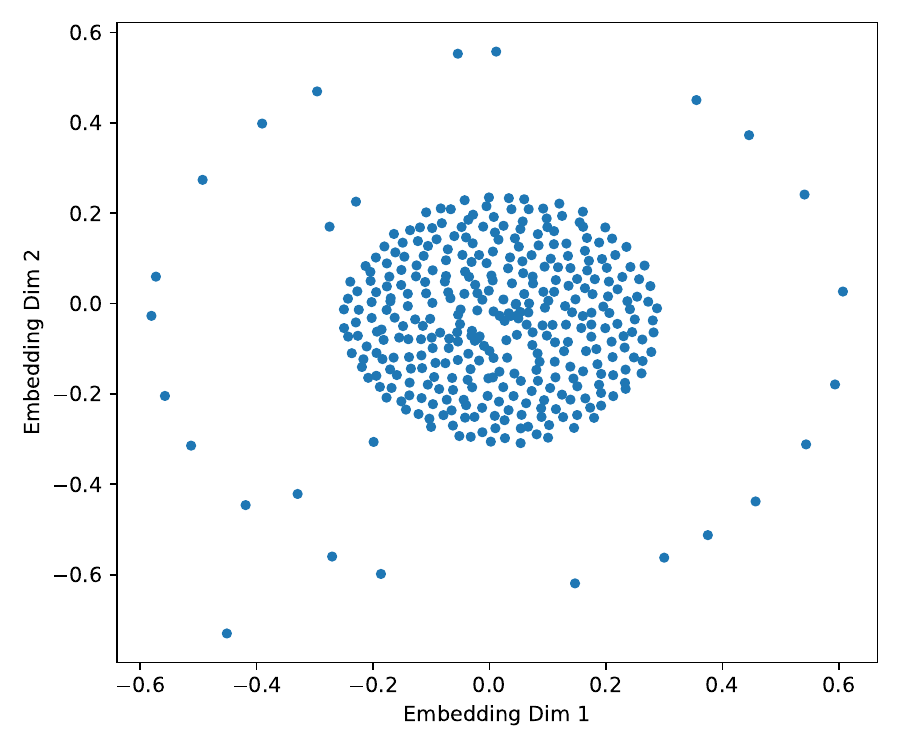}
        \caption{Mean-GWMDS}
        \label{fig:eld_next}
    \end{subfigure}
    \vspace{0.3cm}
    \begin{subfigure}[t]{0.45\textwidth}
        \centering
        \includegraphics[width=\textwidth]{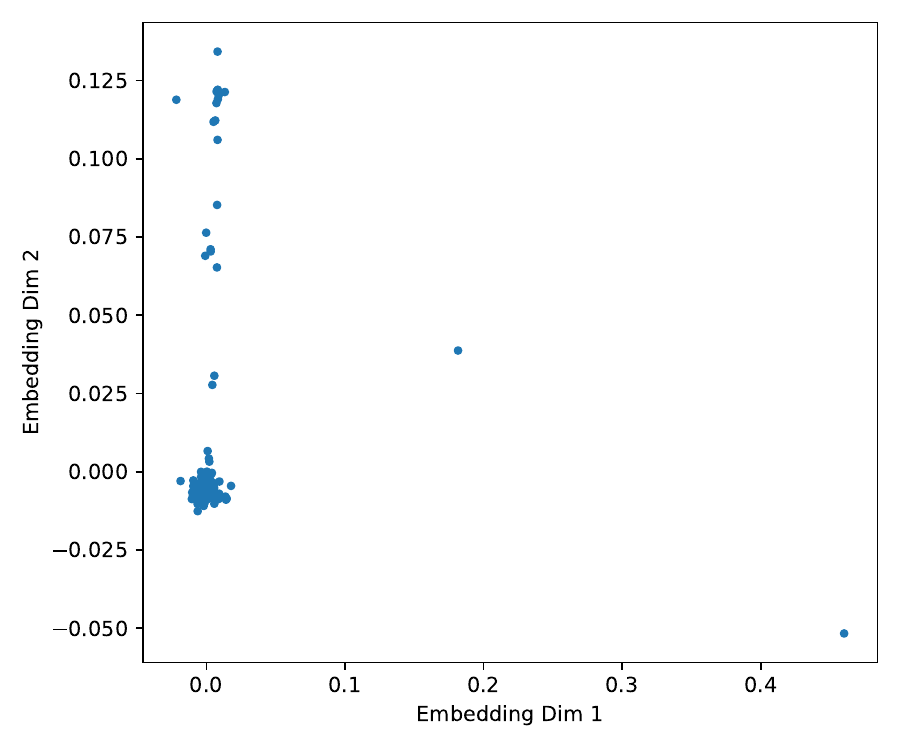}
        \caption{MCCA}
        \label{fig:eld_mcca}
    \end{subfigure}
    \hfill
    \begin{subfigure}[t]{0.45\textwidth}
        \centering
        \includegraphics[width=\textwidth]{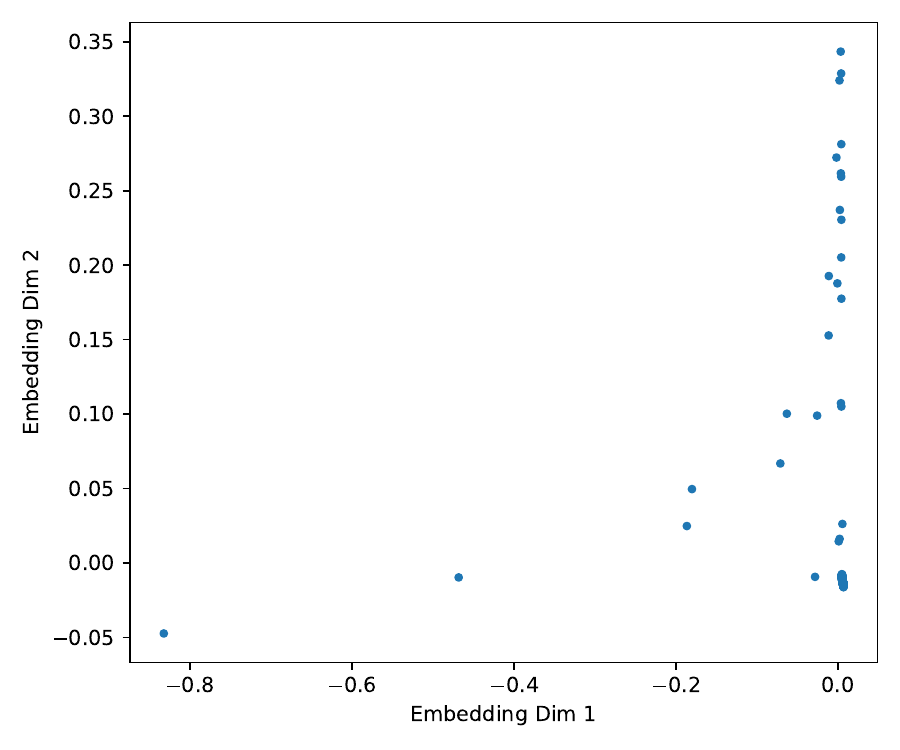}
        \caption{MVMDS}
        \label{fig:eld_mvmmds}
    \end{subfigure}

    \caption{Embeddings on the Electricity Load Diagrams dataset obtained with different multi-view methods. Multi-GWMDS and Mean-GWMDS use Euclidean distances and are contrasted with Euclidean and correlation-based baselines.}
    \label{fig:eld_comparison}
\end{figure}

Figure~\ref{fig:eld_comparison} provides a qualitative comparison of embeddings obtained using Multi-GWMDS, Mean-GWMDS, MVMDS, and MCCA on the Electricity Load Diagrams dataset. The Multi-GWMDS embedding exhibits a coherent and well-organized geometric structure, reflecting the explicit alignment of relational information across views through view-dependent optimal transport couplings. The resulting configuration preserves global relationships while maintaining local consistency, indicating an effective integration of heterogeneous temporal views.

The Mean-GWMDS embedding also yields a structured and visually coherent representation, capturing the overall relational organization of the data through the averaging of view-specific distance structures. However, when compared to Multi-GWMDS, the resulting geometry appears more concentrated and less expressive, suggesting that simple averaging may attenuate view-specific relational variations present in heterogeneous temporal data.

In contrast, the MVMDS embedding shows noticeable geometric distortions. Although some local groupings are preserved, the global structure appears less stable, indicating sensitivity to view aggregation through distance averaging and limitations in handling heterogeneous relational patterns.

The MCCA embedding further highlights the impact of linear assumptions. By operating exclusively on correlations in the feature space, MCCA produces a representation with reduced geometric coherence, failing to adequately capture the nonlinear relational structures present in the data.

Overall, this visual comparison reinforces the quantitative results reported earlier: while Mean-GWMDS provides a strong and stable baseline, explicitly modeling view-dependent relational alignments via optimal transport enables Multi-GWMDS to better preserve the intrinsic structure of real-world multi-view data when compared to Euclidean and correlation-based baselines.

\section{Conclusions and future directions} \label{sec:conclusion}
In this work, we investigated geometry-aware strategies for multi-view representation learning grounded in Gromov-Wasserstein optimal transport, with the goal of integrating heterogeneous views through their intrinsic relational structure rather than feature-level correspondences. We introduced two complementary aggregation mechanisms: a distance averaging strategy, denoted as Mean-GWMDS, which captures consensus geometry across views, and a selection-based strategy, denoted as Multi-GWMDS, which identifies a representative structure via cross-view consistency. These strategies were evaluated under Euclidean and geodesic dissimilarities, highlighting the ability of the proposed framework to adapt to both linear and nonlinear geometric settings. Extensive experiments on synthetic manifolds and a large-scale real-world electricity consumption dataset demonstrate that the proposed methods consistently improve robustness to view heterogeneity and nonlinear distortions when compared to classical multi-view baselines relying on Euclidean or correlation-based assumptions. Beyond the results presented here, several natural extensions emerge from this framework, including the manipulation of view weights $\lambda_v$ to prioritize specific views in a task-driven manner, the integration of heterogeneous dissimilarities across views, such as combining Euclidean and geodesic distances within a single multi-view setting, and the development of clustering-oriented formulations that exploit the semi-relaxed structure to learn representative geometric prototypes, positioning geometry-aware and semi-relaxed optimal transport as a flexible and principled foundation for future advances in multi-view learning.
More broadly, this work highlights the potential of optimal transport as a unifying geometric framework for extending manifold learning to multi-view and heterogeneous settings, where classical feature-based assumptions no longer hold.







\bibliographystyle{elsarticle-num} 
\bibliography{sn-bibliography} 

@article{memoli2011gromov,
  title="{Gromov--Wasserstein distances and the metric approach to object matching}",
  author={M{\'e}moli, Facundo},
  journal={Foundations of Computational Mathematics},
  volume={11},
  pages={417--487},
  year={2011},
  publisher={Springer}
}

@article{
van2024distributional,
title="{Distributional Reduction: Unifying Dimensionality Reduction and Clustering with Gromov-Wasserstein}",
author={{H. Van Assel, C{\'e}dric Vincent-Cuaz, Nicolas Courty, R{\'e}mi Flamary, Pascal Frossard, Titouan Vayer}},
journal={Transactions on Machine Learning Research},
issn={2835-8856},
year={2025},
url={https://openreview.net/forum?id=cllm6SS354},
note={}
}

@article{montesuma2023recent,
  title={Recent advances in optimal transport for machine learning},
  author={Montesuma, Eduardo Fernandes and Mboula, Fred Maurice Ngole and Souloumiac, Antoine},
  journal={IEEE Transactions on Pattern Analysis and Machine Intelligence},
  year={2024},
  publisher={IEEE}
}

@article{peyre2019computational,
  title="{Computational optimal transport: With applications to data science}",
  author={Peyr{\'e}, Gabriel and Cuturi, Marco},
  journal={Foundations and Trends{\textregistered} in Machine Learning},
  volume={11},
  number={5-6},
  pages={355--607},
  year={2019},
  publisher={Now Publishers, Inc.}
}

@inproceedings{clark2024generalized,
  title="{Generalized dimension reduction using semi-relaxed Gromov-Wasserstein distance}",
  author={Clark, Ranthony A and Needham, Tom and Weighill, Thomas},
  booktitle={Proceedings of the AAAI Conference on Artificial Intelligence},
  volume={39},
  number={15},
  pages={16082--16090},
  year={2025}
}

@article{kerdoncuff2021sampled,
  title="{Sampled Gromov-Wasserstein}",
  author={Kerdoncuff, Tanguy and Emonet, R{\'e}mi and Sebban, Marc},
  journal={Machine Learning},
  volume={110},
  number={8},
  pages={2151--2186},
  year={2021},
  publisher={Springer}
}

@inproceedings{Eufrazio,
  title="{A dimensionality reduction technique based on the Gromov-Wasserstein distance}",
  author={Eufrazio, Rafael Pereira and Montesuma, Eduardo Fernandes and Cavalcante, Charles Casimiro},
  booktitle={International Conference on Geometric Science of Information},
  pages={111--120},
  year={2025},
  organization={Springer}
}

@article{yu2025review,
  title={A review on multi-view learning},
  author={Yu, Zhiwen and Dong, Ziyang and Yu, Chenchen and Yang, Kaixiang and Fan, Ziwei and Chen, CL Philip},
  journal={Frontiers of Computer Science},
  volume={19},
  number={7},
  pages={197334},
  year={2025},
  publisher={Springer}
}

@article{chowdhury2025deep,
  title={Deep multi-view clustering: A comprehensive survey of the contemporary techniques},
  author={Chowdhury, Anal Roy and Gupta, Avisek and Das, Swagatam},
  journal={Information Fusion},
  pages={103012},
  year={2025},
  publisher={Elsevier}
}

@article{xu2013survey,
  title={A Survey on Multi-view Learning},
  author={Xu, Chang and Tao, Dacheng and Xu, Chao},
  journal={arXiv preprint arXiv:1304.5634},
  year={2013}
}

@misc{electricityloaddiagrams20112014_321,
  author       = {Trindade, Artur},
  title        = {{ElectricityLoadDiagrams20112014}},
  year         = {2015},
  howpublished = {UCI Machine Learning Repository},
  note         = {{DOI}: https://doi.org/10.24432/C58C86}
}

@article{rodosthenous2024multi,
  title={Multi-view data visualisation via manifold learning},
  author={Rodosthenous, Theodoulos and Shahrezaei, Vahid and Evangelou, Marina},
  journal={PeerJ Computer Science},
  volume={10},
  pages={e1993},
  year={2024},
  publisher={PeerJ Inc.}
}

@misc{kelly2023uci,
  title="{The UCI machine learning repository}",
  author={Kelly, Markelle and Longjohn, Rachel and Nottingham, Kolby},
  year={2023}
}

@incollection{hotelling1992relations,
  title="{Relations between two sets of variates}",
  author={Hotelling, Harold},
  booktitle={Breakthroughs in Statistics: Methodology and Distribution},
  pages={162--190},
  year={1992},
  publisher={Springer}
}

@article{kettenring1971canonical,
  title={Canonical analysis of several sets of variables},
  author={Kettenring, Jon R},
  journal={Biometrika},
  volume={58},
  number={3},
  pages={433--451},
  year={1971},
  publisher={Oxford University Press}
}

@inproceedings{bai2017multidimensional,
  title={Multidimensional scaling on multiple input distance matrices},
  author={Bai, Song and Bai, Xiang and Latecki, Longin Jan and Tian, Qi},
  booktitle={Proceedings of the AAAI Conference on Artificial Intelligence},
  volume={31},
  pages={1281-1287},
  year={2017}
  }

@inproceedings{peyre2016gromov,
  title={Gromov-wasserstein averaging of kernel and distance matrices},
  author={Peyr{\'e}, Gabriel and Cuturi, Marco and Solomon, Justin},
  booktitle={International conference on machine learning},
  pages={2664--2672},
  year={2016},
  organization={PMLR}
}

@article{shi2020approach,
  title={An approach of electrical load profile analysis based on time series data mining},
  author={Shi, Ying and Yu, Tao and Liu, Qianjin and Zhu, Hanxin and Li, Fusheng and Wu, Yaxiong},
  journal={IEEE Access},
  volume={8},
  pages={209915--209925},
  year={2020},
  publisher={IEEE}
}

@article{qin2025survey,
  title={A survey on representation learning for multi-view data},
  author={Qin, Yalan and Zhang, Xinpeng and Yu, Shui and Feng, Guorui},
  journal={Neural Networks},
  volume={181},
  pages={106842},
  year={2025},
  publisher={Elsevier}
}

@article{salter2017latent,
  title={Latent space models for multiview network data},
  author={Salter-Townshend, Michael and McCormick, Tyler H},
  journal={The Annals of Applied Statistics},
  volume={11},
  number={3},
  pages={1217},
  year={2017}
}

@article{sun2013survey,
  title={A survey of multi-view machine learning},
  author={Sun, Shiliang},
  journal={Neural Computing and Applications},
  volume={23},
  number={7},
  pages={2031--2038},
  year={2013},
  publisher={Springer}
}

@article{lin2023multi,
  title={Multi-view clustering via optimal transport algorithm},
  author={Lin, Renjie and Du, Shide and Wang, Shiping and Guo, Wenzhong},
  journal={Knowledge-Based Systems},
  volume={279},
  pages={110954},
  year={2023},
  publisher={Elsevier}
}

@article{zhang2024learning,
  title={Learning common semantics via optimal transport for contrastive multi-view clustering},
  author={Zhang, Qian and Zhang, Lin and Song, Ran and Cong, Runmin and Liu, Yonghuai and Zhang, Wei},
  journal={IEEE Transactions on Image Processing},
  year={2024},
  publisher={IEEE}
}




\end{document}